# Buggy rule diagnosis for combined steps through final answer evaluation in stepwise tasks

Gerben van der Hoek[1][0009-0004-0932-3065], Johan Jeuring[1][0000-0001-5645-7681],

and Rogier Bos[1][0000-0003-2017-9792]

[1] Utrecht University PO Box 85.170, 3508 AD Utrecht, the Netherlands
g.vanderhoek@uu.nl

**Abstract.** Many intelligent tutoring systems can support a student in solving a stepwise task. When a student combines several steps in one step, the number of possible paths connecting consecutive inputs may be very large. This combinatorial explosion makes error diagnosis hard. Using a final answer to diagnose a combination of steps can mitigate the combinatorial explosion, because there are generally fewer possible (erroneous) final answers than (erroneous) solution paths. An intermediate input for a task can be diagnosed by automatically completing it according to the task solution strategy and diagnosing this solution. This study explores the potential of automated error diagnosis based on a final answer. We investigate the design of a service that provides a buggy rule diagnosis when a student combines several steps. To validate the approach, we apply the service to an existing dataset (n=1939) of unique student steps when solving quadratic equations, which could not be diagnosed by a buggy rule service that tries to connect consecutive inputs with a single rule. Results show that final answer evaluation can diagnose 29,4% of these steps. Moreover, a comparison of the generated diagnoses with teacher diagnoses on a subset (n=115) shows that the diagnoses align in 97% of the cases. These results can be considered a basis for further exploration of the approach.

## 1 Introduction

In education, we see a rapid growth of computer-aided assessment through intelligent tutoring systems (ITSs) (Sangwin, 2013). For instance, many textbooks offer online tutoring environments in addition to written tasks. Online tutoring fulfills a demand for efficient, specific, and timely feedback on learning processes. ITSs track students' progress in learning trajectories and support educators in responding adequately to hurdles students may encounter (Nwana, 1990). Recently, the use of generative AI in education has increased (Rane, 2023; Troilo et al., 2024; Yoon et al., 2024). Learning opportunities can be created through finding useful prompts to converse with such systems. These dialogues benefit from the unpredictability that results from the stochastic processes that govern generative AI. However, when formatively assessing students' work there is a danger that generative AI misses the point. For instance, Jia et al. (2024) show that the quality of hints generated by AI was significantly inferior to the quality of human hints. In this paper we study possibilities to improve feedback using more classical AI-methods.

An important function of providing hints or feedback is to prevent a student from persistently making an error (Narciss, 2008). Such feedback can be based on a diagnosis of the student input. A widely used approach to diagnosing such input is model tracing (MT) (Anderson et al., 1995). In MT the steps a student is supposed to take to solve a stepwise task are modeled by production rules. The rules transform an expression in the same way a student should, to correctly solve the task. In such a model of the task so-called buggy rules can be incorporated to mimic erroneous student steps (VanLehn & Brown, 1980). As such an ITS can reason about a task in much the same way a student could.

The part of an ITS that can reason with expert knowledge is called a domain reasoner. An example of a state-of-the-art framework for stepwise MT domain reasoners is the IDEAS framework. The core of the IDEAS framework is a domain-specific language (DSL) for defining strategies for stepwise problem solving. IDEAS MT domain reasoners are used in many applications for domains such as functional programming, rewriting logic expressions, hypotheses testing, polynomial equations, and linear and exponential extrapolation (Gerdes et al., 2012, 2017; Heeren et al., 2010; Heeren & Jeuring, 2014, 2020; Tacoma et al., 2020; Van der Hoek et al., 2024).

Most domain reasoners check for erroneous input by trying to apply a single buggy rule. Applying combinations of rules can easily lead to combinatorial explosion, making computations hard to manage (Ohlsson & Mitrovic, 2007). The IDEAS domain reasoner currently also checks a single rule through its buggy rule service (Heeren & Jeuring, 2014). Diagnosing errors by checking only a single buggy rule is a significant restriction. For instance, logs show that 88,9% of the unique erroneous steps could not be diagnosed by the IDEAS service in a study of Bokhove & Drijvers (2012) on feedback in an algebra tutor for quadratic equations. This brings us to the following research question. How can we design a service that provides a buggy rule diagnosis when a student combines several, sometimes all, steps of an expected strategy?

To answer this research question, we introduce an approach to diagnosing an intermediate student step through its final answer. The key to this approach is that an intermediate expression can be viewed as a (sub) task, which has a solution. For instance, when solving a linear equation an intermediate expression can have the form $a_1 \cdot x + b_1 = a_2 \cdot x + b_2$, which has the solution $x = \frac{b_2 - b_1}{a_1 - a_2}$. If a student applies a rule to the expression transforming it into an equivalent expression, the solution remains the same. However, if a student applies a buggy rule, the resulting expression is not equivalent, and the solution changes. Hence, the final answer of an intermediate expression carries information about buggy rules that might have been applied.

We call the collection of all possible (erroneous) intermediate expressions given certain rules and buggy rules, the search space of a task. Below we argue that analyzing the collection of unique final answers of intermediate expressions instead of all expressions in the search space could mitigate combinatorial explosion associated with multistep MT (Ohlsson & Mitrovic, 2007). Each (possibly erroneous) MT solution path has only one final answer, so the number of final answers is less than or equal to the number of intermediate expressions in the solution paths. Moreover, often different orders of applying rules yield the same result. For instance, if a student erroneously solves the linear equation $3x + 1 = x + 2$ by subtracting instead of dividing in the last step, i.e., $2x = 1 \rightarrow x = -1$ then all solution paths the student could have used prior to the last step, yield this same erroneous final answer. Generally, a final answer is reachable through different solution paths. Hence final answer diagnoses could mitigate combinatorial explosion. We call our approach to diagnosing erroneous student input using final answers model backtracking (MBT). The main goal of MBT is to deduce applied buggy rules from an intermediate student input, through the final answer obtained by applying the solution strategy to the input.

In this paper, we first elaborate on literature involving various aspects of MBT. We then illustrate the various aspects involved in MBT through examples. After which we show how teachers evaluate MBT-diagnoses generated for steps in the logs of the Bokhove & Drijvers (2012) study.

## 2 Theoretical framework

Below we elaborate on the MT approach and buggy rules, and discuss the IDEAS framework for MT domain reasoners and its implementation of strategies. Combinatorial explosion can occur in MT, but normal forms can be used to mitigate it. An instance of a normal form is a final answer.

Model backtracking is a rule-based approach that can be classified as a model tracing variant. In model tracing, expert knowledge is embedded in a domain reasoner through production rules (Anderson et al., 1995). Tasks that can be solved by transforming a starting expression into a solution expression can be implemented through model tracing. Here, the implemented production rules transform the expression the same way a student should. For example, if a student solves an equation of the type $a_1x + b_1 = b_2$, one rule would be "$a_1x + b_1 = b_2 \rightarrow a_1x = b_2 - b_1$". If a student deviates from what is expected by the model in the domain reasoner, an error or deviation can be flagged. Buggy rules can be incorporated to provide specific error diagnoses. An example of a buggy rule for solving linear equations of the above type is: "$a_1x + b_1 = b_2 \rightarrow a_1x = b_2 + b_1$".

Expert knowledge can also be embedded in a domain reasoner through constraint-based modeling (CBM) (Mitrovic, 2012). CBM embeds expert knowledge in the form of pairs of constraints. The first constraint, the so-called relevance condition, determines whether a situation is such that the second constraint, the satisfaction condition, should be checked. In the example above of solving: $a_1x + b_1 = b_2$ a relevance condition might be: *"the student input is of the form: $ax = b$"* and the corresponding satisfaction condition: "$b/a = (b_2 - b_1)/a_1$". If the satisfaction condition is violated an error can be flagged. CBM does not implement buggy rules, but detects errors through violated constraints.

An MT domain reasoner can be implemented using the IDEAS framework (Heeren & Jeuring, 2014). A solution procedure for a task is modeled using a DSL-expression (Heeren & Jeuring, 2017). IDEAS' DSL consists of combinations of rules similar to regular expressions where the symbols are the basic production rules. The DSL comprises several combinators that construct DSL expressions from other DSL expressions. Table 1 gives a list of common combinators. We call a DSL expression a strategy because it models a solution strategy.

**Table 1.** Selection of strategy combinators in the IDEAS DSL

| Combinator | Notation | Description |
| --- | --- | --- |
| Composition | `s .*. t` | Apply strategy `t` after strategy `s` |
| Choice | `s <\|> t` | Apply strategy `s` or strategy `t` |
| Many | `many s` | Apply strategy `s` zero or more times |
| Repeat | `repeat s` | Repeat strategy `s` until no longer applicable |

A DSL expression can be applied to a task to generate the search space for the task and solution procedure. Applying a combinatorial structure, such as a strategy, to a starting task is subject to combinatorial explosion (Ohlsson & Mitrovic, 2007). To avoid combinatorial explosion, the IDEAS diagnose service tries to connect two consecutive student inputs by a single rule. Using a single buggy rule to diagnose student errors is restrictive (Bokhove & Drijvers 2012). We discuss measures to mitigate combinatorial explosion when using multiple rules to connect consecutive student inputs.

To mitigate combinatorial explosion, we reduce the search space by identifying similar objects and taking a single example of each class of similar objects. It is common to take equality of normal forms to determine similar objects (Heeren & Jeuring, 2009; Sangwin, 2013). We further abstract from normal forms of objects in a domain $D$ by defining a function of type $D \rightarrow D$ that satisfies properties such as idempotency and semantic preservation. If all objects in a knowledge domain have a desired, or final, state (e.g., the solution or final answer), then mapping an object to its desired state gives a normal form. As an example, consider the domain of solving quadratic equations. The set of objects $D$ for this domain consists of quadratic equations or pairs of linear equations. The desired final state of an object $q \in D$ is $\{x = z_q^1,\ x = z_q^2\}$ where $z_q^1$, and $z_q^2$ are the solutions of $q$. The function $f: q \rightarrow \{x = z_q^1,\ x = z_q^2\}$ defines a normal form.

When analyzing a normal form of a student input instead of the student input itself, we lose information about the way the student reasons about the task. The extent of congruence between the way a person reasons about a task, and the way knowledge is implemented in a domain reasoner is called cognitive fidelity (Mitrovic et al., 2003; Ohlsson & Mitrovic, 2007). Generally, an MT approach has a high cognitive fidelity. Arguably, an MT domain reasoner that only analyzes a normal form of a student input, such as the final answer, has lower cognitive fidelity than an MT domain reasoner that analyzes the literal student input.

If the goal of analyzing student input is to provide feedback about specific errors or bugs, then the loss of information and cognitive fidelity caused by the use of a normal form might lead to decreasing specificity of feedback. Shute (2008) defines the specificity of feedback as the complexity of the information the feedback contains. In the case of feedback on bugs, it can vary from information about the correctness of a response to elaborated feedback about a specific error a student made. Elaborated feedback can serve to reduce the cognitive load on a student's working memory while executing a task (Sweller et al., 1998). The studies Shute (2008) analyzed show that complex feedback can support a student in the learning process by alleviating cognitive load, but can also prevent students from learning to find and correct errors themselves. As such, moderate feedback complexity might be effective. Below we shall find that MBT diagnoses support feedback of moderate complexity.

In the next section, we describe the design of a buggy rule service based on the final answer normal form.

## 3 Design

We elaborate on aspects of our design through examples that showcase the various aspects of MBT. The aspects involved are (1) defining a buggy strategy, (2) final answer evaluation, (3) intermediate search space reduction, (4) generating diagnoses and (5) disambiguating tasks. Note that buggy rule application has already been implemented with success in many systems (Gertner & VanLehn, 2000; Jones & VanLehn, 1992; Matsuda et al., 2007). Below, we are therefore interested in providing a buggy rule diagnosis only when steps are combined.

### 3.1 Defining a buggy strategy

Consider the following situation: a student is tasked with stepwise reducing an expression $a_1 + a_2 + \cdots + a_n$ by adding together a pair of adjacent terms at each step, until a single term is left. Suppose students make two types of errors when completing such a task: (a) subtracting instead of adding, and (b) forgetting the first term of a pair of adjacent terms. Using the DSL, we model this strategy and its buggy rules by: `buggyStrategy = repeat ("add adjacent"<|>"subtract adjacent"<|>"forget first")` This DSL-expression generates all possible final answers when applied to a starting task. These final answers can be used to diagnose student input as we will see below.

### 3.2 Final answer evaluation

MBT uses final answer evaluation, which has both strengths and weaknesses. Consider the example described in Table 2. Applying the three different buggy rules to $2 = x - 3$ yields the same final answer, namely $x = -1$. Any of the three buggy rules can be detected in a final answer diagnosis by a single buggy rule: $ax + b = c \rightarrow ax + b = -c$. Thus, final answer evaluation allows us to define a buggy strategy compactly and subsequently mitigate combinatorial explosion. However, this mitigation comes at a price: decreased cognitive fidelity (Ohlsson & Mitrovic, 2007), and decreased diagnosis specificity; because the three different errors cannot be distinguished. This means we should be cautious with providing feedback based on a final answer diagnosis. However, even in the single buggy rule diagnosis of "add 3 on the right, subtract it on the left" in Table 2, the step might also have been a combination of "drop the – in −3" and a mathematically correct step. We further discuss diagnoses when we discuss generating them and in the section on the evaluation of MBT.

Table 2. Three examples of buggy rule application

| Start | | $2 = x - 3$ | |
|---|---|---|---|
| Buggy rule | Switch $x$ and 2 | Add 3 on the right, subtract it on the left | Drop – in $-3$ |
| Result | $x = 2 - 3$ | $2 - 3 = x$ | $2\ =\ x\ + 3$ |

### 3.3    Intermediate search space reduction

Intermediate search space reduction is the main technique for mitigating combinatorial explosion. To illustrate the technique, we calculate in how many ways the expression $1 + \cdots + 6$ can be reduced using the buggyStrategy above. In the first step of buggyStrategy there are five ways to choose a pair of adjacent terms. For each pair of terms, we can apply any of the three rules from the buggyStrategy yielding: $5 \cdot 3$ possibilities for the first step. Continuing, we find four locations to apply any of the three rules for all of the $5 \cdot 3$ results from the previous step, yielding $5 \cdot 3 \cdot 4 \cdot 3 = 3^2 \cdot 5 \cdot 4$ paths with two steps. Reducing the entire expression in this fashion we obtain $3^5 \cdot 5! = 29160$ paths. The number of all intermediate expressions is the sum of the expressions in each step: $\sum_{i=1}^{5} 3^i \cdot \frac{5!}{(5-i)!} = 40695$. Now, each path ends with a final answer, where multiple paths can have the same final answer. If we define a final answer as the final expression in each path, then the search space of final answers is smaller than the search space of all intermediate expressions by a factor: $\frac{29160}{40695} \approx .72$.

This shows that in this way final answer diagnosis reduces the required amount of computation, but not yet enough. Looking at the expression $1 + \cdots + 6$ we find that the minimal possible result using buggyStrat is $1 - (2 + \cdots + 6) = -19$ and the maximal possible result is $1 + \cdots + 6\ = 21$ . Therefore, the number of unique final answers is at most 41. Consequently, the search space of unique final answers is smaller than the search space of all intermediate expressions by a factor $\frac{41}{40695} \approx 0.001$.

Applying buggyStrategy as a DSL-expression to $1 + \cdots + 6$ yields a list of 29160 final answers, one for each path. Reducing this list of 29160 final answers into a set of at most 41 unique final answers is probably more expensive than searching through the list of 29160 final answers. Therefore, we apply the reduction intermediately, at strategic moments during the computation of the search space. When the number of expressions in the search space exceeds a predetermined limit, we delete all but one of the identical expressions that have the same remaining strategy. This reduces the number of paths needed to compute the unique final answers. In our results section we will provide examples of the computing time reduction achieved by intermediate search space reduction.

### 3.4 Generating diagnoses

To generate diagnoses, we use tuples of the form $(x, I)$ where $x$ is an expression and $I$ is a set of sets that contain identifiers for buggy rules. Whenever a buggy rule is applied to $x$ the rule adds an identifier to each set in $I$. Suppose now we are given a student input $q$. We then compute the final answer normal form associated with $q$, say $q'$. We search the unique calculated final answer tuples for a tuple $(x, I)$ such that $x = q'$. Then $I$ is the MBT diagnosis for the input $q$; it has the following form: $I = \{\{id_1^1, \cdots, id_{n_1}^1\}, \cdots, \{id_1^k, \cdots, id_{n_k}^k\}\}$, where the sets $J_i = \{id_1^i, \cdots, id_{n_i}^i\}$ are 'as small as possible' see Figure 1 and Table 3.

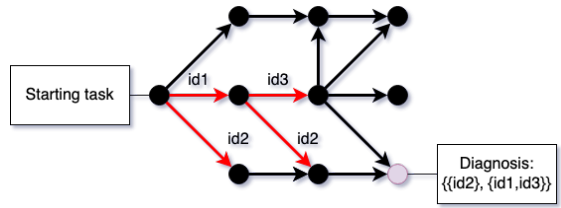

**Fig. 1.** Schematic representation of an MBT diagnosis

Each $J_i$ is a possible way to reach the final answer $x$ using the smallest number of buggy rules $id_1^i, \cdots, id_{n_i}^i$ and a number of non-buggy rules. We call such a set $J_i$ an MBT-diagnosis alternative. Because the $J_i$ are as short as possible we avoid listing repeated buggy rules that can be simplified into smaller combinations of rules (for instance: negating the same term three times). Notice that in Figure 1 $\{id1, id2\}$ is not part of de diagnosis, because the student input can already be reached (explained) by just $\{id2\}$.

**Table 3.** Two examples of MBT-diagnoses for quadratic equations

| Example 1 | Example 2 |
|---|---|
| Input1 = $[4(x + 6)^2 + 3 = 39]$ | Input1 = $[2x - 5 = 0,\ x - 5 = -2x + 4]$ |
| Input2 = $[x = 9]$ | Input2 = $\left[x = \frac{5}{2}, x = 9\right]$ |
| MBT-diagnosis: | MBT-diagnosis: |
| $\{J_1 = \{$ "negate a term" | $\{J_1 = \{$ "forget divide"$\}$ |
| , "forget an equation"$\}\}$ | $, J_2 = \{$ "negate a term"$\}\}$ |

### 3.5 Disambiguating tasks

Disambiguating tasks is an optional technique to increase final answer diagnosis accuracy. To illustrate it we return to our example of reducing an expression $a_1 + a_2 + \cdots + a_n$ in a stepwise fashion with the buggyStrategy. Consider the starting tasks $ex1 : 1 + 2 + 3 + 4$, and $ex2 : 1 + 9 + 3 + 4$. By calculating the number of paths produced by buggyStrategy for both expressions we find $3^3 \cdot 3! = 162$ paths. However, when calculating the number of unique final answers for both expressions we find that $ex1$ has 18 unique final answers and $ex2$ has 23 unique final answers. Since both expressions have the same number of paths this must mean that $ex1$ has more paths that coincide than $ex2$. When two coinciding paths include different combinations of buggy rules, they lead to MBT-diagnosis alternatives. This shows that we can adjust the parameters of the starting task to reduce the number of alternatives in MBT diagnoses. Consequently, one way to enhance MBT-diagnosis accuracy is to select the parameters of a starting task such that the number of unique final answers is large. Disambiguating starting tasks in this manner is possible, for instance, for randomized tasks, where the choice of the parameters of the starting task is inconsequential aside from perhaps a specified number domain.

The MBT technique comprises the five aspects above: (1) define a buggy strategy, (2) evaluate to a final answer, (3) reduce search space at intermediate steps, (4) generate diagnoses, (5) disambiguate tasks. In the next section we evaluate the MBT-technique by comparing MBT diagnoses to teacher diagnosis for the case of polynomial equations.

## 4     Evaluation of MBT diagnoses

In this section, we elaborate on the participants, the instrument, data collection, data analysis, and the results of an evaluation of MBT-diagnoses for buggy rules in solving quadratic equations.

*Participants:* Our data holds data from the research by Bokhove & Drijvers (2012). Their intervention took place in 2010 in fifteen 12th-grade classes from nine Dutch secondary schools ($n = 324$), with students of age 17 or 18. The schools were spread across the country and showed a variation in school size and pedagogical and religious backgrounds. The participating classes were of pre-university level. 43 % of the participants were female and 57 % were male.

*Instrument:* Students were presented a task to solve a quadratic equation using the strategies that are part of the Dutch mathematics curriculum (Bokhove & Drijvers, 2012). The tasks were embedded in the digital math environment (Bokhove, 2017) equipped with an IDEAS domain reasoner. A student could enter a solution in a stepwise fashion to receive feedback on the use of their strategy and on possible errors.

*Data collection:* We used data on solving quadratic equations. The data consists of 5493 diagnose requests for mathematically incorrect steps. We removed duplicate requests, which resulted in 2284 steps. Of these unique steps 254 could be diagnosed by IDEAS buggy service (11,1 %), leaving 2030 steps that could not be diagnosed. Of these 2030 steps, our MBT buggy service could diagnose 662 requests. 91of these diagnoses consisted of a single buggy rule application. These buggy rules were not implemented in IDEAS, but if they had been, IDEAS would have given the same diagnosis. This leaves us with 571 requests diagnosable by MBT out of 1939 requests that could not be diagnosed by IDEAS (29,4%). These 571 steps comprise the dataset for further analyses.

*Data analysis:* For analysis, a random sample of 115 steps out of the 571 MBT-diagnosable steps was taken. The first and third author who are also experienced mathematics teachers carried out an analysis of the steps by coding them independently. To this end, the third author compiled a codebook of buggy rules. The codebook consisted of four buggy rules for situations in which a student forgot an equation, five buggy rules for situations with an error with a minus sign, and five miscellaneous buggy rules, such as approximation of a square root or writing a fraction upside down. The teachers could also submit a code "no idea" when they could not identify the student error.

After coding, the results of the authors/teachers were compared. The proportion of matching codes was equal to .55. Because in each item several steps were combined the errors in a number of steps were explainable in multiple ways, resulting in disagreement between the teachers. However, each of the four buggy rules describing a forgotten equation could be grouped under a single code: "forget an equation". Furthermore, the five buggy rules describing an error with a minus sign could be grouped under the code: "negate a term". Using these new codes, the proportion of matching codes was .97, where the teachers only disagreed on three items. After a discussion, the teachers fully agreed about a coding of the steps using the codes: "forget an equation", "negate a term", and the five codes for the miscellaneous buggy rules.

*Results:* MBT diagnoses were compared to the teacher coding of the steps. Table 3, in the generating diagnoses section, shows two examples of MBT-diagnoses that aligned with the teacher coding. On three items the MBT-diagnosis did not align with the teacher coding. For two of these items the teachers did not agree after grouping the buggy rules but prior to the discussion. Both teachers could not identify the student error in the third item. However, they found that the MBT-diagnosis did not apply. This brings the proportion of aligned diagnoses to $\frac{112}{115} \approx .97$. With this proportion we can compute a Wilson score confidence interval for $\alpha = .1$: $[.94; .99]$. The lower bound of this interval is a lower bound for the probability of an aligned MBT diagnosis for $\alpha = .05$. This shows that a lower bound for the probability of aligned MBT-diagnoses is .94. The Wilson score interval for the proportion of misaligned diagnosis for $\alpha = .1$ equals $[.01; .06]$. Consequently, an upper bound for the probability of unaligned diagnoses or misdiagnoses with $\alpha = .05$ is .06. This shows that the 1939 requests that could not be diagnosed by IDEAS can be partitioned in three segments, (1) at least $.94 \cdot 29,4\% \approx 27,6\%$ receives a correct diagnosis, (2) at most $.06 \cdot 29,4\% \approx 1,8\%$ receives a misdiagnosis, and (3) the remaining $100\% - 29,4\% = 70.6\%$ receives no diagnosis. In the next section we discuss the various results.

## 5     Evaluation of computational efficiency

Computational efficiency of the reduction of the search space technique is evaluated in two ways. Firstly, by determining the average computing time for a sample of steps from the data of Bokhove & Drijvers (2012), and secondly by a hypothetical investigation. All computing times below were determined on a Macbook Air with an Apple M1 chip. For the first evaluation a sample of 50 steps was taken from the dataset of 2030 steps that IDEAS could not diagnose. The average computing time for a possible diagnosis without using reduction of the search space was 3.64 seconds, where the longest time was 17.23 seconds. With reduction, the average computing time was .35 seconds with a longest time of 2.11 seconds. In this case, the longest computing time was decreased by more than eight times.

For the second evaluation, we define a hypothetical buggy strategy similar to the example in the design section. To do so, we first define rules $R_i$ that map two adjacent terms of an expression to a random integer $k$ with $|k| < 10^6$, such that $R_i(a + b) \neq R_j(a + b); i \neq j$. Furthermore, we define $\varepsilon(k)$ as a sum of $k$ random integers. This is an expression that can be reduced analogously to the example in the design section. Now, our strategy is defined by: `hypoStrat` $n$ = `repeat(`$R_1$`<|>`$\cdots$`<|>`$R_n$`)`. Table 4 shows the computing times in seconds (s) required for calculating a list of unique final answers by applying `hypoStrat` $n$ to $\varepsilon(k)$. The calculation was completed through two modes, either normal, where the intermediate reduction of the search space was not applied, or reduction, where reduction was applied after each iteration of the repeat combinator. A dash (-) indicates the calculation was terminated due to overflow. The relative increase in computational efficiency is greater when either the expression or the strategy is 'larger'. Furthermore, the intermediate search space reduction mitigates combinatorial explosion, but also suffers from it.

**Table 4.** Computing time of unique final answers by applying `hypoStrat` $n$ to $\varepsilon(k)$

| `hypoStrat` $n$ | mode | $\varepsilon(6)$ | $\varepsilon(7)$ | $\varepsilon(8)$ | $\varepsilon(9)$ |
|---|---|---|---|---|---|
| $n = 2$ | normal | .03 (s) | .33 (s) | 4.84 (s) | - |
| | reduction | .01 (s) | .11 (s) | .76 (s) | 6.32 (s) |
| $n = 3$ | normal | .18 (s) | 3.54 (s) | - | - |
| | reduction | .11 (s) | 1.09 (s) | 13.39 (s) | - |
| $n = 4$ | normal | .76 (s) | - | - | - |
| | reduction | .43 (s) | 6.33 (s) | - | - |

## 6   Discussion of the MBT technique

Here we discuss the evaluation of the MBT technique. We have found that for solving quadratic equations MBT can diagnose 29,4% of the unique requests that IDEAS could not diagnose. Moreover, the probability of alignment with a teacher diagnosis, is at least .94 ($\alpha = .05$). MBT was applied to an existing dataset, which implies that the tasks in the dataset were not modified to increase diagnosis accuracy. Hence the actual lower bound for the probability of aligned diagnosis could be higher. To realize this .94 lower bound, several similar errors were grouped together, potentially decreasing feedback specificity (Shute, 2008). To illustrate the decrease in feedback specificity we look at the earlier example of solving: $2 = x - 3$ for $x$. Dropping the minus in $-3$, yields: $2 = x + 3$, and moving the $-3$ to the other side of the equation without changing the sign yields: $2 - 3 = x$. Both errors yield an equation with the same solution. Therefore, these errors are indistinguishable for final answer evaluation. This phenomenon was also the cause of the disagreement between the teachers during the initial coding process. It shows that errors that produce equivalent expressions should be grouped when diagnosing combined steps.

Considering the above, a feedback message responding to a student error should take all the potential errors a student could have made into account. Hence, the errors in solving quadratic equations that were grouped under "negate a term" should receive feedback along the lines of: *"It seems you have made a mistake with a sign; this could be caused by moving a term to the other side of the equation"*. Can such MBT-generated feedback be helpful to students? Shute (2008) found that elaborate feedback helps a student in the learning process by alleviating cognitive load (Sweller et al., 1998), however, students also should have opportunities to learn to find and correct errors themselves. In that respect, a feedback message such as above could be helpful. It alleviates cognitive load by providing a suggestion about what type of error was made, but still leaves an opportunity for a student to find and correct the error.

In prior research a feedback scheme where feedback is provided while leaving opportunities for self-guidance was implemented in a tutor for linear and exponential extrapolation. Here, an MBT-service provided error-specific feedback and selected suitable follow-up tasks. In 21 out of 25 episodes where a student improved without worked examples or direct instruction, the improvement was preceded by either error-specific feedback or a follow-up task, in four cases improvement was preceded by right or wrong feedback (Van der Hoek et al., 2024b).

The previous two paragraphs show that a student could benefit from MBT diagnoses in various situations. How generalizable is the MBT technique? IDEAS' domain-specific language (Heeren & Jeuring, 2017) is domain independent. Therefore, it can be used in any domain where the objects have a desired state or final answer normal form. In such domains grouping buggy rules producing equivalent expressions reduces the number of buggy rules in a buggy strategy mitigating combinatorial explosion. This combinatorial explosion can be further mitigated by intermediate search space reduction. However, combinatorial explosion is merely mitigated and not removed. Table 4 shows that there are bounds to the computational efficiency of MBT.

CBM embeds expert knowledge without the risk of combinatorial explosion (Ohlsson & Mitrovic, 2007). What does MBT offer above CBM? Consider the example $3x - 1 = x + 2$ and suppose a student only enters: $x = 2$. Here, the MBT diagnosis is $\{\{\text{"negate a term"}, \text{"reverse divide"}\}\}$. Now, imagine general constraints on a solution of the form $x = b$ such that violation of the constraints provides a similar diagnosis, without calculating erroneous final answers. Providing such constraints will probably be difficult. Here we have a situation where the student does not provide an intermediate step. Furthermore, a solution of the form $x = b$, is one-dimensional and has little internal structure to formulate constraints for detecting these kinds of errors. Mitrovic & Ohlsson (2006) state that it is very hard for both CBM and MT to perform a substantial diagnosis in this situation. However, MBT can provide a diagnosis in this situation.

Whether or not MBT can also provide students with meaningful advice is still an open question at this point. Nonetheless, above we argued that MBT-generated feedback might help students. Moreover, we showed the feasibility of the approach because we achieved an average computing time of .35 seconds for a diagnosis. In this study, we found that MBT diagnoses are feasible for solving quadratic equations, and in prior research, we found feasibility for linear extrapolation, and exponential extrapolation (Van der Hoek et al., 2024a, 2024b). These task types are of comparable complexity to other upper secondary school mathematics tasks. Therefore, we expect MBT to be generalizable to at least such domains.

## 7     Conclusion and future work

In this section, we answer our research question on the design of a service that provides a buggy rule diagnosis when a student combines several, possibly all, steps of an expected strategy. For the case of quadratic equations, this research question is answered by the application of MBT to this domain. The MBT service could diagnose 29,4% of the steps where a single buggy rule service could not. Furthermore, a lower bound for the probability of a successful diagnosis is .94. This lower bound can be further increased by disambiguating tasks.

We expect MBT to be generalizable to at least upper-secondary school mathematics tasks. To support this, we provided arguments for the generalizability of the MBT approach. For tasks involving complex expressions and strategies, the MBT approach might not be feasible anymore with limited resources. In complex domains, embedding objects in an egraph (Willsey et al., 2021) could extend computational feasibility. An egraph is a data structure for efficiently calculating and storing expressions resulting from production rule applications.

In the near future, we plan to further study whether students can benefit from MBT feedback through classroom experiments with an environment for quadratic equations. In the more distant future, we plan to implement MBT with IDEAS domain reasoners for other domains. The MBT software will be published using an open-source license, making it broadly available.

**Acknowledgments** We thank Paul Drijvers for his contributions to the study as a whole. Furthermore, we especially thank Bastiaan Heeren, who laid the groundwork for this study.